\documentclass{article}
\usepackage{spconf,amsmath}
\usepackage{graphicx}
\usepackage{amssymb}
\usepackage{amsmath}
\usepackage{booktabs}
\usepackage{multirow}
\usepackage{tabularx}
\usepackage{array}
\usepackage{hyperref} 

\title{Sketch2Symm: Symmetry-Aware Sketch-to-Shape Generation via Semantic Bridging}
\name{Yan Zhou$^{1}$, Mingji Li$^{2}$\sthanks{© 2025 IEEE. Personal use of this material is permitted. Permission from IEEE must be obtained for all other uses, in any current or future media, including reprinting/republishing this material for advertising or promotional purposes, creating new collective works, for resale or redistribution to servers or lists, or reuse of any copyrighted component of this work in other works.}, Xiantao Zeng$^{2}$, Jie Lin$^{1}$, Yuexia Zhou$^{1}$}
\address{$^{1}$School of Electronic Information Engineering, Foshan University, Guangdong, China\\
$^{2}$School of Computer Science and Artificial Intelligence, Foshan University, Guangdong, China}

\begin{document}
\ninept
\maketitle

\begin{abstract}
Sketch-based 3D reconstruction remains a challenging task due to the abstract and sparse nature of sketch inputs, which often lack sufficient semantic and geometric information. To address this, we propose Sketch2Symm, a two-stage generation method that produces geometrically consistent 3D shapes from sketches. Our approach introduces semantic bridging via sketch-to-image translation to enrich sparse sketch representations, and incorporates symmetry constraints as geometric priors to leverage the structural regularity commonly found in everyday objects. Experiments on mainstream sketch datasets demonstrate that our method achieves superior performance compared to existing sketch-based reconstruction methods in terms of Chamfer Distance, Earth Mover's Distance, and F-Score, verifying the effectiveness of the proposed semantic bridging and symmetry-aware design.
\end{abstract}


\section{Introduction}
\label{sec:intro}
In recent years, the rapid advancement of deep learning has significantly accelerated progress in 3D shape reconstruction from 2D images, opening up broad application prospects. Accurately generating 3D shapes from 2D images enhances realism and precision, which is crucial for industries such as virtual reality, robotics, and manufacturing. Effectively leveraging the rich visual information contained in 2D images to reconstruct 3D shapes is key to improving the efficiency and quality of digital representations. However, in real-world scenarios, sketches are often a more common and accessible form of expression. Therefore, extending the research focus from image-driven to sketch-driven 3D reconstruction holds substantial importance.

Compared with images that contain rich visual cues such as color, texture, and shading, sketches inherently suffer from sparsity and abstraction. To address these challenges, existing studies have explored various strategies, including style normalization~\cite{zhong2020towards}, data augmentation~\cite{gao2022sketchsampler}, and additional priors such as viewpoint estimation~\cite{zhang2021sketch2model,zheng2023locally} and multi-view sketches~\cite{zhou2023ga,chen2024sketch2nerf,ying2025sketchsplat}. Some approaches also leverage learned 3D shape priors~\cite{sanghi2023sketch}. Representative works include: SketchSampler~\cite{gao2022sketchsampler}, which translates sketches into more informative 2D representations via image-to-image networks; PASTA~\cite{lee2025pasta}, which integrates sketches with text descriptions in vision–language models; S3D~\cite{song2025s3d}, which introduces style alignment loss and augmentation; and SketchDream~\cite{liu2024sketchdream}, which employs a sketch-based multi-view image diffusion model with depth guidance. Additionally, image translation has advanced rapidly, with CoCosNet~\cite{CoCosNet} emerging as a powerful and extensible framework that has inspired numerous task-specific studies. Some works introduce contrastive learning to enhance cross-domain invariant feature modeling~\cite{MCL-Net}, while others explore unpaired exemplar-guided translation to improve domain transfer~\cite{dsi2i}. These advances in image translation have also provided insights and inspiration for sketch-to-shape generation tasks.

Despite extensive exploration, existing methods still show limited generalization to sketches of varying styles and often fail to fully exploit their sparse information. Meanwhile, many everyday objects exhibit strong symmetry~\cite{li2022symmnerf}, which humans frequently rely on in perception and imagination. For example, one can infer the missing half of a chair from only one side. However, current approaches rarely model symmetry explicitly, and the inherent sparsity and abstraction of sketches make it difficult to leverage symmetry effectively. Therefore, translating sketches into images as a semantic bridge becomes crucial for both enriching sparse representations and enabling effective symmetry modeling.

To address this, we propose a sketch-based two-stage 3D reconstruction method. In the first stage, we perform sketch-to-image translation, extending CoCosNet~\cite{CoCosNet} for this specific task. This allows sketches to be effectively transformed into semantically rich images, serving as more informative and structured intermediate representations that bridge the domain gap and improve the accuracy of subsequent 3D prediction. In the second stage, the intermediate images are fed into a geometry-aware network for 3D shape reconstruction. During training, we incorporate an explicit symmetry constraint as a geometric prior, which compensates for the lack of 3D structural cues in sketches and encourages the generation of more complete and regular 3D structures. 

The main contributions of this paper can be summarized as: 1) We propose a semantic bridging strategy via sketch-to-image translation, which enriches sketch representations and facilitates more effective 3D reconstruction. 2) A geometric symmetry constraint is incorporated during training as an explicit prior to encourage object-level structural regularity, improving the completeness and plausibility of reconstructed shapes. 3) We perform extensive experiments on public sketch datasets, demonstrating superior performance in reconstruction accuracy, generalization, and robustness compared to representative sketch-based 3D generation methods.

\section{Method}
\label{sec:format}

We propose Sketch2Symm, a two-stage method for sketch-based 3D shape reconstruction. The training procedure of both stages is illustrated in Fig.~\ref{fig1}. The first stage (Section~\ref{sec2.1}) employs a cross-domain image translation network to convert sparse sketches into semantically enriched images, thereby enhancing the geometric cues available for reconstruction. The second stage (Section~\ref{sec2.2}) incorporates a symmetry-based loss into an image-to-point-cloud generation pipeline, encouraging structural regularity and detail preservation. The two stages are trained independently to address the semantic sparsity and structural ambiguity inherent in sketches. During inference (Section~\ref{sec2.3}), a single sketch is processed sequentially through both stages to produce a structurally consistent 3D point cloud.

\subsection{Stage 1: Sketch-to-Image Translation}
\label{sec2.1}
The abstractness and sparsity of sketches limit the available semantic information, which poses significant challenges for 3D reconstruction. To mitigate this issue, we first synthesize shape-consistent images from input sketches. This step enriches the sketch representation by leveraging the richer semantic capacity of natural images, thereby compensating for the limited information in sketches. 

\begin{figure}[t]  
\centering
\includegraphics[width=\columnwidth]{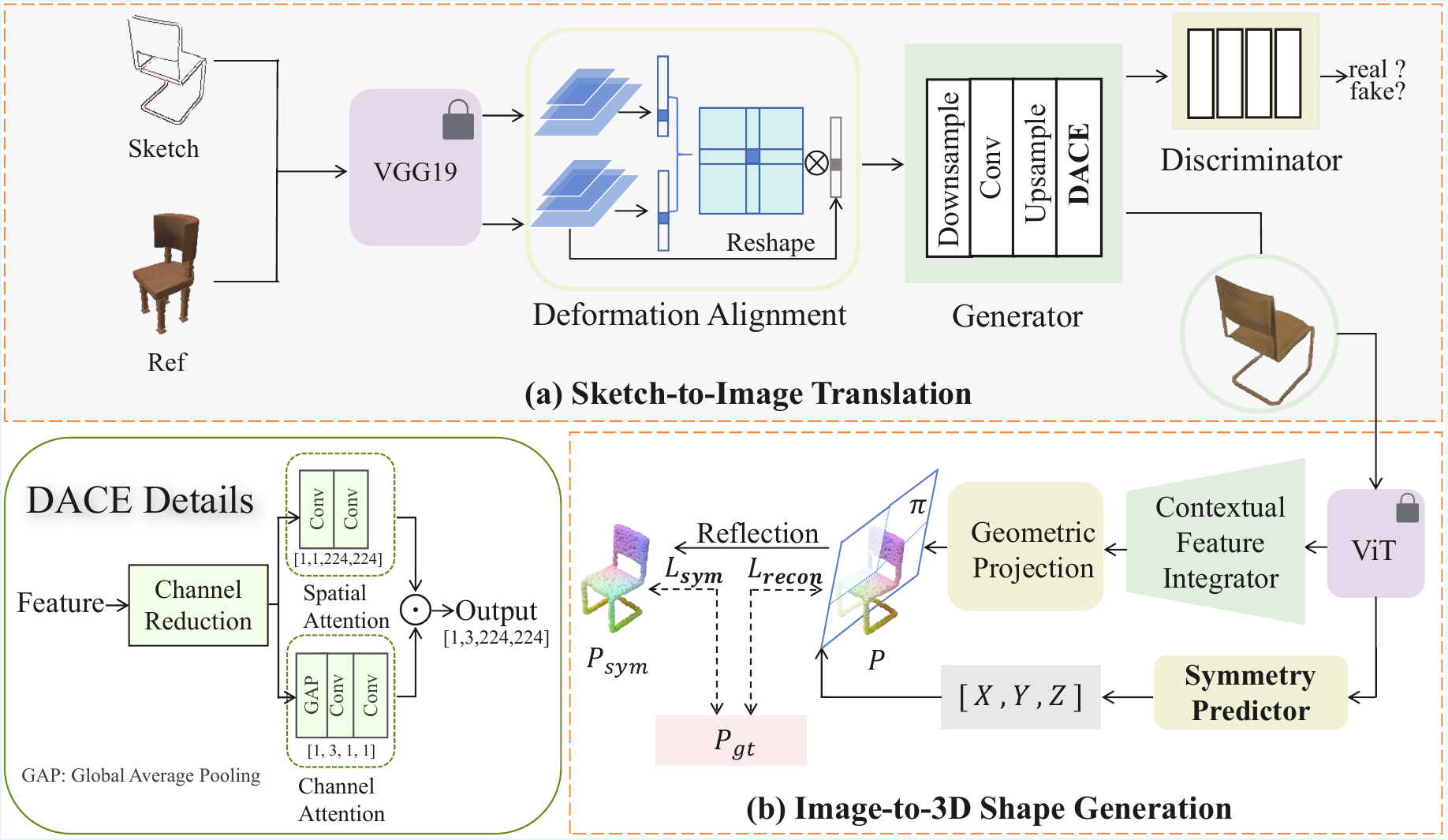} 
\caption{The two-stage training pipeline of Sketch2Symm. (a) illustrates the training process of Stage 1 from left to right. (b) shows the training process of Stage 2 from right to left. The bottom left corner shows the details of DACE.} 
\label{fig1}
\end{figure}

In the first stage, the processing pipeline begins with multi-scale feature extraction using a pre-trained VGG-19 model~\cite{brock2018large}, which processes both input sketches and reference images to obtain hierarchical feature representations. These features are then fed into the Deformation Alignment Network, as illustrated in Fig.~\ref{fig1}(a), which establishes pixel-wise correspondences between cross-domain inputs through deep feature correlation and cosine similarity computation. This cross-modal alignment deforms the geometric structure of the reference image to match the sketch, thereby guiding the image synthesis process in the downstream adversarial network.

In the Deformation Alignment Network, we establish cross-domain deep correspondences between sketch and reference image. We first construct a shared semantic space $s$ within the latent feature domain to align the representations of input sketches and reference images. Specifically, an input sketch $x$ is mapped to its feature representation $x_s$, while a reference image $y$ is mapped to its feature representation $y_s$. In the shared space $s$, we employ cosine similarity as the similarity metric. By maximizing the cosine similarity between $x_s$ and $y_s$, we encourage the two representations to be directionally consistent and semantically aligned. The cosine similarity is defined as:
\vspace{-3pt}
\begin{equation}
\cos(\theta) = \frac{x_s^T y_s}{\|x_s\|_2 \cdot \|y_s\|_2},
\end{equation}
and our optimization objective is formulated as:
\vspace{-3pt}
\begin{equation}
\max_{\theta_x, \theta_y} \frac{x_s^T y_s}{\|x_s\|_2 \cdot \|y_s\|_2},
\end{equation}
which enforces semantic alignment between sketches and reference images in the shared domain.

Due to the inherent lack of chromatic and textural information in sketch inputs, which requires full reliance on reference images for color and texture synthesis, we introduce the Dual-Attention Color Enhancement (DACE) module to enable adaptive feature refinement, as illustrated in the generator part of Fig.~\ref{fig1}(a). DACE is located at the final layer of the generator and employs a collaborative dual-attention mechanism through two dedicated branches. Firstly, to adapt the feature vectors to the input of DACE, we apply a Channel Reduction defined as:
\vspace{-3pt}
\begin{equation}
\max(1, {in\_channels} // factor).
\end{equation}
We set the factor to 4 with the aim of effectively reducing the computational cost by about 50\% while still retaining sufficient representational capacity. The spatial attention branch learns positional importance via two cascaded convolutions, emphasizing geometric regions that require enhanced coloration. The channel attention branch first applies global average pooling, followed by two convolutional layers, to capture inter-channel dependencies and adaptively adjust the enhancement strength of each color component through a channel importance learning mechanism. The optimized attention features from both branches are then fused via element-wise multiplication with broadcasting. This hierarchical attention fusion markedly improves color fidelity and visual realism, while alleviating texture misalignment issues in cross-modal synthesis tasks. 

\subsection{Stage 2: Image-to-3D Shape Generation}\label{sec2.2}
After completing the generation from sketches to images in the first stage, the second stage aims to further reconstruct 3D shapes with geometric consistency. To enhance structural plausibility, we introduce a symmetry constraint during the generation process. We adopt point cloud as the 3D representation, as its coordinate-based form naturally supports the application of symmetry through geometric transformations.

Among existing image-to-3D reconstruction methods, we adopt RGB2Point~\cite{RGB2Point} as the baseline for the second stage, due to its simple encoder-decoder architecture and coordinate-based output. It uses a pre-trained Vision Transformer (ViT)~\cite{dosovitskiy2020image} to extract semantic features from the input image, which are then enhanced by a Contextual Feature Integrator and mapped to 3D point coordinates via Geometric Projection. This direct coordinate prediction facilitates the incorporation of explicit geometric constraint. To further improve structural plausibility, we introduce a symmetry constraint during training as an additional regularization term. By defining a reflective mapping in 3D space, each predicted point is paired with its mirrored counterpart, encouraging the model to better capture the inherent symmetry present in many objects.

To explicitly model the symmetrical structures commonly found in real-world objects, we introduce the symmetry constraint during the generation process. Specifically, as illustrated in Fig.~\ref{fig1}(b), we assume that there exists a symmetric plane $\pi$, which is parameterized by the unit normal vector $\mathbf{n} = [X, Y, Z] \in \mathbb{R}^3$ and the offset $d \in \mathbb{R}$, and its geometric definition is:
\vspace{-3pt}
\begin{equation}
\pi: \mathbf{n}^\top \mathbf{x} + d = 0.
\end{equation}

\renewcommand{\arraystretch}{0.8} 
\begin{table*}[htbp]
\centering
\caption{Quantitative comparison of CD on thirteen categories. CD values are multiplied by $10^{3}$.}
\label{tab1}
\begin{tabular}{lccccccc} 
\toprule
\multirow{2}{*}{Method} & \multicolumn{7}{c}{Chamfer Distance $\downarrow$}\\
\cmidrule(lr){2-8} 
 & Car & Sofa & Airplane & Bench & Display & Chair & Table \\
\midrule
DISN~\cite{xu2019disn} & 7.90& 17.65& 11.59 & 12.97& 16.63 & 15.17& 25.86 \\
Sketch2Model~\cite{zhang2021sketch2model}  & 15.26 & 42.35 & 22.94 & 23.31 & 24.07 & 61.96 & 21.87 \\
Sketch2Mesh~\cite{guillard2021sketch2mesh}    & 11.48& 25.73& 9.29& 9.01& 15.67& 16.83& 17.81\\
Deep3DSketch~\cite{chen2023deep3dsketch} & 12.57& 43.96& 23.41& 23.54& 23.31& 61.27& 20.84\\
Deep3DSketch-im~\cite{chen2024deep3dsketch} & 7.42& 17.76& 11.72& 13.61& 16.90& 15.35& 25.72\\
\textbf{Ours}    & \textbf{2.10}& \textbf{4.70}& \textbf{1.50}& \textbf{5.00}& \textbf{5.70}& \textbf{5.00}& \textbf{7.50}\\
\midrule\midrule 
\multirow{2}{*}{Method} & \multicolumn{7}{c}{Chamfer Distance $\downarrow$}  \\
\cmidrule(lr){2-8}
 & Telephone & Cabinet & Loudspeaker & Watercraft & Lamp & Rifle & \multicolumn{1}{|c}{Average} \\
\midrule
DISN~\cite{xu2019disn} & 8.79& 16.08& 18.67& 16.24& 40.30& 8.03& \multicolumn{1}{|c}{16.53} \\
Sketch2Model~\cite{zhang2021sketch2model}  & 18.82 & 18.67 & 20.73 & 15.72 & 60.34 & 19.00 & \multicolumn{1}{|c}{28.08} \\
Sketch2Mesh~\cite{guillard2021sketch2mesh}    & 17.62& 20.44& 12.06& 8.99& 33.29& 8.87& \multicolumn{1}{|c}{15.93} \\
Deep3DSketch~\cite{chen2023deep3dsketch} & 16.11& 18.36& 22.23& 15.25& 56.41& 19.30& \multicolumn{1}{|c}{27.43} \\
Deep3DSketch-im~\cite{chen2024deep3dsketch} & 8.66& 15.85& 19.04& 16.18& 30.38& 7.95& \multicolumn{1}{|c}{15.89} \\
\textbf{Ours}    & \textbf{2.60}& \textbf{8.50}& \textbf{10.80}& \textbf{3.40}& \textbf{7.90}& \textbf{1.70}& \multicolumn{1}{|c}{\textbf{5.11}} \\
\bottomrule
\end{tabular}
\end{table*}
\renewcommand{\arraystretch}{1.0} 
\vspace{-10pt}

To predict the symmetry plane from input images, we introduce a Symmetry Predictor module that takes the extracted image features as input and outputs the normal vector components through a Multi-layer Perceptron with ReLU activations. The predicted normal vector is then used to construct a 3×3 reflection matrix $\mathbf{R}$ according to the formula $\mathbf{R} = \mathbf{I} - 2\mathbf{n}\mathbf{n}^\top$, where $\mathbf{I}$ is the identity matrix. For each 3D point $\mathbf{p}_i$ generated by the network, we construct its symmetric counterpart $\mathbf{p}_i^*$ about the symmetry plane using the reflection transformation:
\vspace{-3pt}
\begin{equation}
\mathbf{p}_i^* = \mathbf{p}_i - 2 (\mathbf{n}^\top \mathbf{p}_i + d) \mathbf{n} = \mathbf{R} \mathbf{p}_i.
\end{equation}

As shown in Fig.~\ref{fig1}(b), our network simultaneously generates both the original point cloud $P$ and its symmetric counterpart $P_{\text{sym}}$ by applying the reflection matrix to all generated points. The symmetry constraint is enforced through a dual supervision strategy:
\vspace{-3pt} 
\begin{equation}
\mathcal{L}_{3D} = \mathcal{L}_{\text{recon}}(P, P_{\text{gt}}) + \mathcal{L}_{\text{sym}}(P_{\text{sym}}, P_{\text{gt}}),
\end{equation}

where $\mathcal{L}_{\text{recon}}$ represents the original RGB2Point~\cite{RGB2Point} reconstruction loss, and $\mathcal{L}_{\text{sym}}$ adopts the same reconstruction loss formulation as $\mathcal{L}_{\text{recon}}$ to measure the geometric similarity between the symmetric point cloud and ground-truth. This dual supervision encourages the network to generate point clouds that not only match the ground-truth directly, but also maintain consistency when reflected across the predicted symmetry plane, effectively enforcing geometric symmetry in the generated 3D structures.

\subsection{Inference}\label{sec2.3}
In the first inference stage, the same frozen VGG-19 model~\cite{brock2018large} used during training is adopted for feature extraction. Unlike training, no user-provided reference images are required at inference; instead, the system uses fixed reference images. This design is motivated by the use of sketch–reference pairs during training to improve generalization, whereas inference focuses on enriching sketches with image-level color and texture.

In the second stage, a frozen pre-trained ViT~\cite{dosovitskiy2020image} extracts features from the synthesized image, which are then fed into the Contextual Feature Integrator and Geometric Projection Module for point cloud reconstruction. Unlike the training phase, symmetric plane prediction is not required during inference. By sequentially executing these two stages, the method produces a complete and structurally consistent point cloud.

\section{Experiments}
\subsection{Experimental Settings}
\subsubsection{Datasets} To support our two-stage training pipeline, we employ the ShapeNet-Synthetic~\cite{zhang2021sketch2model} sketch dataset and the image dataset by Xu et al.~\cite{xu2019disn}. Both datasets are derived from the ShapeNet Core dataset~\cite{chang2015shapenet} with corresponding categories and shapes. For the first training stage, we use the ShapeNet-Synthetic dataset and Xu et al.'s dataset. For the second training stage, we use the rendered images and their corresponding 3D shapes from the ShapeNet Core dataset. For performance evaluation, we utilize both the ShapeNet-Synthetic dataset and the ShapeNet-Sketch~\cite{zhang2021sketch2model} dataset, which contains hand-drawn sketches derived from the ShapeNet Core dataset.
\subsubsection{Evaluation Metrics} To quantitatively evaluate the quality of the generated 3D point clouds, we employ three widely adopted metrics: Chamfer Distance (CD), Earth Mover's Distance (EMD), and F-Score. F-Score is computed with a threshold of 0.01.

\subsubsection{Implementation Details}
All training and evaluation are conducted on a single 24GB NVIDIA RTX 3090Ti GPU, with the model configured to generate point clouds containing 2048 points. In the training of Stage 1, we use the Adam optimizer with learning rates of $1 \times 10^{-4}$ for the generator and $4 \times 10^{-4}$ for the discriminator, applying spectral normalization to ensure training stability. In the training of Stage 2, we use the Adam optimizer with an initial learning rate of $5 \times 10^{-4}$, applying learning rate decay with a factor of 0.7 when the validation loss plateaus, and using gradient clipping with a maximum norm of 5.0.

\subsection{Quantitative Analysis}
In Table~\ref{tab1}, our method achieves the lowest CD values across all thirteen object categories in the ShapeNet-Synthetic~\cite{zhang2021sketch2model} dataset, demonstrating superior performance compared to existing representative methods. In addition to CD, we further evaluate EMD and F-score in three classic categories: Chair, Car, and Airplane, and compare the results among three representative methods. The results are reported in the upper part of Table~\ref{tab2}. Our method achieves the lowest EMD scores across all categories, indicating better performance in point cloud alignment and structural consistency.

\subsection{Qualitative Analysis}
\begin{figure}
\includegraphics[width=\columnwidth]{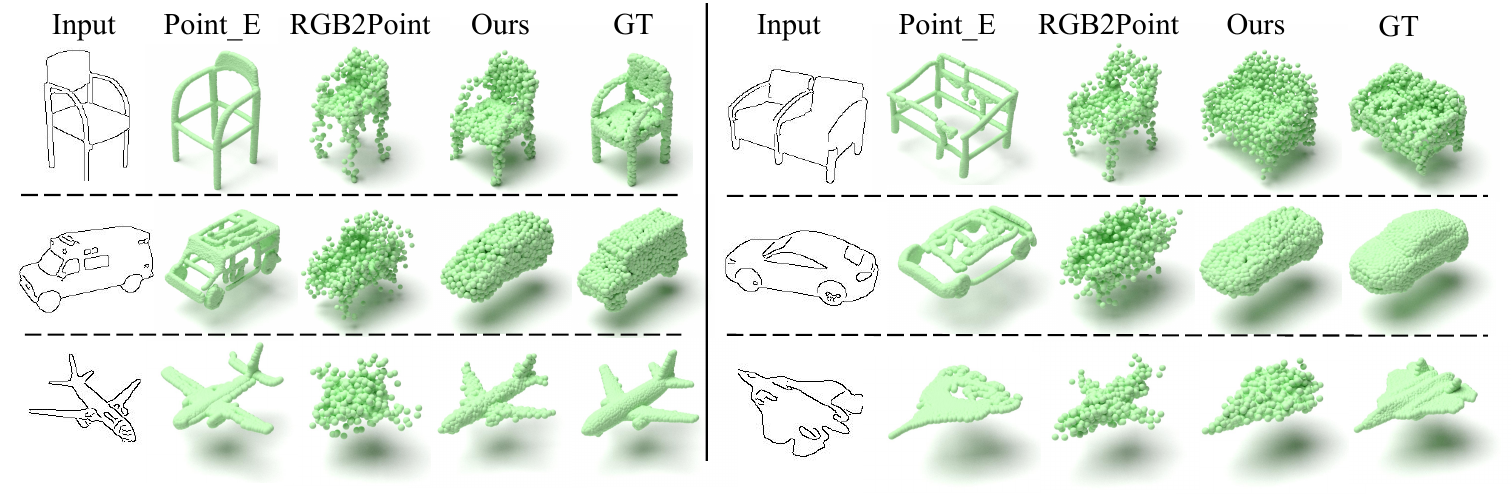}
\caption{Visualization of qualitative comparison on the ShapeNet-Synthetic dataset using synthetic sketches.} \label{fig2}
\end{figure}

\begin{figure}
\includegraphics[width=\columnwidth]{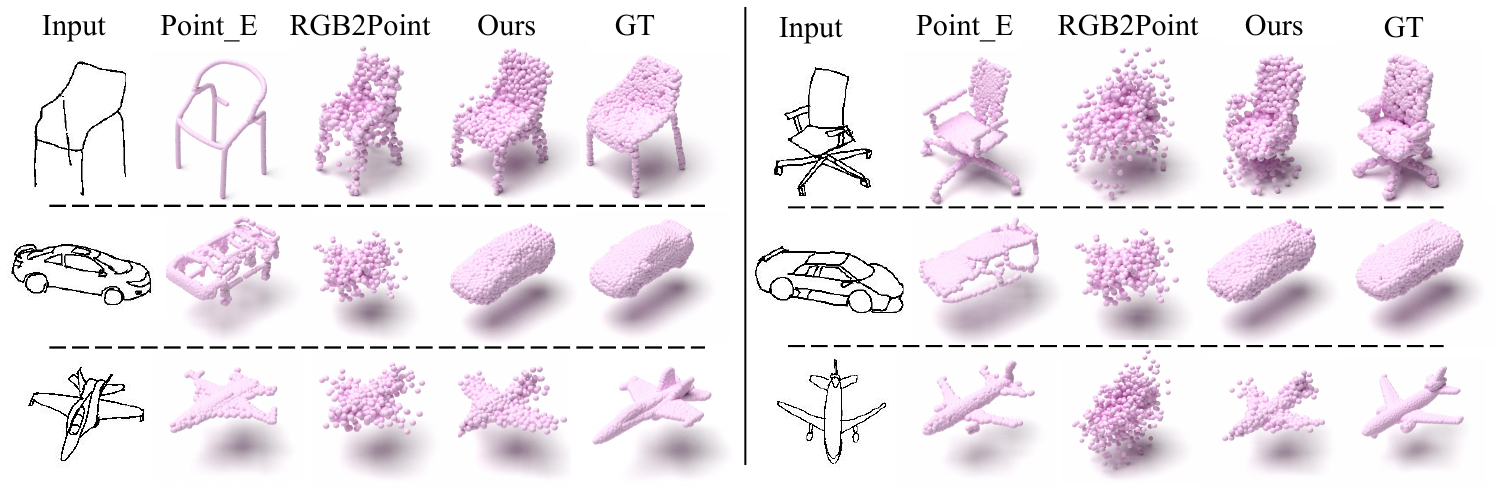}
\caption{Visualization of qualitative comparison on the ShapeNet-Sketch dataset using hand-drawn sketches.} \label{fig3}
\end{figure}

To qualitatively evaluate reconstruction quality, we focus on three categories: Chair, Car, and Airplane, conducting detailed visual assessments. we visualize the generated point clouds in Fig.~\ref{fig2} and Fig.~\ref{fig3}, comparing our method with the representative point cloud reconstruction method Point\_E~\cite{nichol2022point} and the baseline RGB2Point~\cite{RGB2Point}. Fig.~\ref{fig2} uses synthetic sketches from the ShapeNet-Synthetic~\cite{zhang2021sketch2model} dataset, while Fig.~\ref{fig3} uses hand-drawn sketches from the ShapeNet-Sketch~\cite{zhang2021sketch2model} dataset. As demonstrated, our approach demonstrates superior capability in recovering both global structural coherence and fine-grained local details, irrespective of whether the input is a synthetic or hand-drawn sketch.

\subsection{Ablation Studies}
To validate the effectiveness of our proposed components, we conduct ablation experiments on three categories, focusing on the independent contributions of two core components: the intermediate sketch-to-image generation step and the symmetry constraint. The first variant removes the sketch-to-image module to assess the impact of semantically enriched images on 3D reconstruction; the second variant excludes the symmetry constraint to evaluate its effect on structural regularity in the generated shapes.

We compare these two ablation variants with our complete method. In the lower of Table~\ref{tab2} reports the results using EMD, and F-Score as evaluation metrics. Removing the sketch-to-image module results in the most significant performance degradation, underscoring its importance. Overall, the complete method achieves the best performance across all metrics, validating the complementary roles of the sketch-to-image conversion and symmetry constraint.

\begin{table}[htbp]
\centering
\caption{Quantitative comparison of EMD and F-Score on three categories. EMD values are multiplied by $10^{-2}$.}
\label{tab2}
\setlength{\tabcolsep}{0.5pt} 

\begin{tabular}{lcccccccc}
\toprule
\multirow{2}{*}{Method} & \multicolumn{4}{c}{EMD$\downarrow$} & \multicolumn{4}{c}{F-Score$\uparrow$} \\
\cmidrule(lr){2-5} \cmidrule(lr){6-9}
& Chair & Car & Airplane & Avg & Chair & Car & Airplane & Avg \\
\midrule
Point-E\cite{nichol2022point}        & 2.94 & 3.58 & 1.73 & 2.75 & 0.35 & 0.19 & 0.53 & 0.36 \\
RGB2Point\cite{RGB2Point}      & 2.20 & 0.85 & 0.45 & 1.17 & 0.53 & 0.91 & \textbf{0.99} & 0.81 \\
\midrule
\textbf{Ours}              & \textbf{1.65} & \textbf{0.27} & \textbf{0.44} & \textbf{0.79} & \textbf{0.95} & \textbf{0.99} & \textbf{0.99} & \textbf{0.98} \\
w/o Sketch-to-Image   & 1.75 & 0.77 & 0.52 & 1.01 & 0.87 & 0.90 & 0.98 & 0.92 \\
w/o Symmetry          & 1.77 & 0.34 & 0.45 & 0.85 & 0.94 & 0.98 & \textbf{0.99} & 0.97 \\
\bottomrule
\end{tabular}
\end{table}
\vspace{-10pt}

\subsection{Comparison with Diffusion-based Methods}

\begin{figure}
\includegraphics[width=\columnwidth]{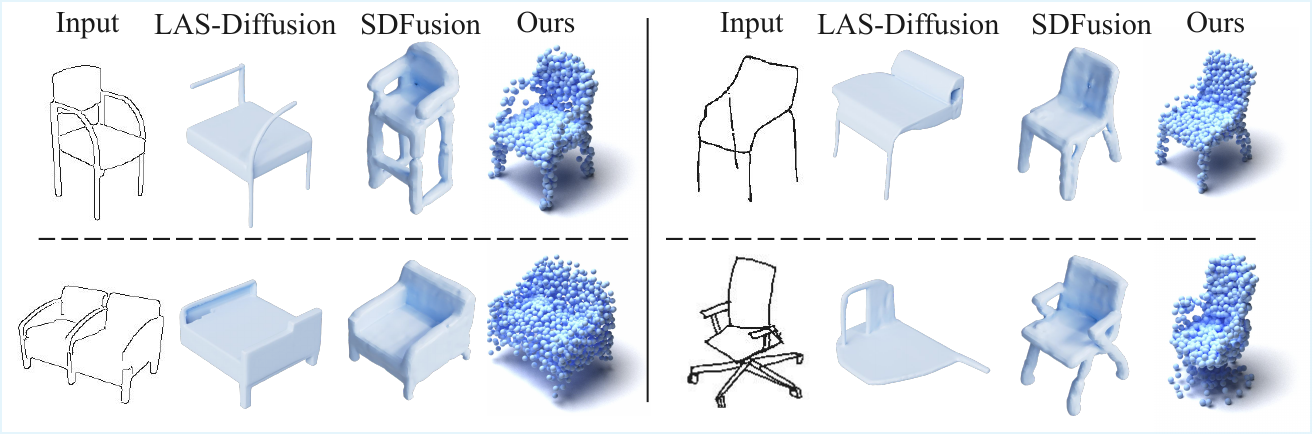}
\caption{Qualitative comparison with diffusion-based methods on synthetic and hand-drawn sketches.} \label{fig4}
\end{figure}

To address the concern that non-diffusion methods might be inferior to diffusion-based approaches, we compare our method with two representative diffusion-based approaches: SDFusion~\cite{cheng2023sdfusion} and LAS-Diffusion~\cite{zheng2023locally}. Both methods are based on mesh generation. As shown in Fig.~\ref{fig4}, our method demonstrates superior visual quality compared to these diffusion-based methods. The left half of Fig.~\ref{fig4} shows results on synthetic sketches, while the right half displays results on hand-drawn sketches.

Table~\ref{tab3} presents a comprehensive comparison in terms of computational efficiency and model complexity. Our method significantly outperforms the diffusion-based approaches in both inference time and parameter count. This comparison demonstrates that our non-diffusion approach not only produces superior visual results but also offers substantial advantages in computational efficiency and resource utilization.

\vspace{-10pt}
\renewcommand{\arraystretch}{0.8} 
\begin{table}[htbp]
\centering
\caption{Computation complexity analysis.}
\label{tab3}
\begin{tabular}{
    l              
    @{\hspace{12pt}} 
    c              
    @{\hspace{12pt}} 
    c              
    @{\hspace{12pt}} 
    c              
    @{\hspace{5pt}} 
}
\toprule
{Method} & {Inference Time} & {Params} \\
\midrule
SDFusion\cite{cheng2023sdfusion} & 7.17s & 1126.34M \\
LAS-Diffusion\cite{zheng2023locally} & 25.99s & 699.94M \\
\textbf{Ours} & \textbf{2.37s} & \textbf{324.69M} \\
\bottomrule
\end{tabular}
\end{table}
\renewcommand{\arraystretch}{1.0} 
\vspace{-10pt}

\section{Conclusion}
In this study, we propose a novel two-stage method for sketch-to-3D shape generation that introduces images as intermediate semantic bridges and employs symmetry-aware geometric reconstruction. Our method shows competitive results compared to state-of-the-art methods on ShapeNet-related datasets. Experiments verify our method's superiority in reconstruction accuracy and structural integrity. This research provides a novel technical pathway for 3D modeling based on non-expert user inputs. Future work will extend our approach to better handle asymmetric objects, broadening applicability to more diverse and complex shapes.


\bibliographystyle{IEEEbib}
\begin{small}
\bibliography{references}    
\end{small}
\end{document}